\documentclass[letterpaper, 10 pt, conference]{ieeeconf} 

\IEEEoverridecommandlockouts                              % Necessario per \thanks e altri comandi
\usepackage{cite}
\usepackage{amsmath,amssymb,amsfonts}
\usepackage{algorithm}
\usepackage{algpseudocode}
\usepackage{textcomp}
\usepackage{etoolbox}
\algrenewcommand\algorithmiccomment[1]{\hfill\(\triangleright\) \textit{\footnotesize#1}}
\usepackage{pdflscape}
\usepackage{graphicx}
\usepackage[dvipsnames,svgnames]{xcolor}
\usepackage{tabularx}
\usepackage{booktabs}
\usepackage{balance}
\usepackage{accents}
\usepackage{array}
\usepackage{url}

\usepackage[caption=false,font=footnotesize]{subfig}

\usepackage[colorlinks=true,linkcolor=black,citecolor=red,urlcolor=black]{hyperref}

\def\BibTeX{{\rm B\kern-.05em{\sc i\kern-.025em b}\kern-.08em
    T\kern-.1667em\lower.7ex\hbox{E}\kern-.125emX}}

\usepackage[acronym,indexonlyfirst,nomain,xindy,toc]{glossaries}
\newacronym{phri}{pHRI}{Physical Human-Robot Interaction}
\newacronym{vf}{VF}{virtual fixture}
\newacronym{lbd}{LbD}{Learning by Demonstration}
\newacronym{lfd}{LfD}{Learning from Demonstration}
\newacronym{kt}{KT}{Kinesthetic Teaching}
\newacronym{ssa}{SSA}{Spatial Sampling Algorithm}
\newacronym{emg}{EMG}{Electromyography}
\newacronym{rd}{R\&D}{Resection and Dissection}
\newacronym{nasatlx}{NASA-TLX}{NASA Task Load Index}
\newacronym{cobot}{cobot}{collaborative robot}
\newacronym{dmp}{DMP}{Dynamic Movement Primitive}
\newacronym{adl}{ADL}{Activities of Daily Living}
\newacronym{mis}{MIS}{Minimally Invasive Surgery}
\newacronym{gvf}{GVF}{Guidance Virtual Fixture}
\newacronym{frvf}{FRVF}{Forbidden Region Virtual Fixture}
\newacronym{dof}{DoF}{degrees of freedom}
\newacronym{rcm}{RCM}{Remote Center of Motion}
\newacronym{ds}{DS}{Dynamical System}
\newacronym{pf}{PF}{Potential Field}
\newacronym{aan}{AAN}{assist-as-needed}
\newacronym{gmm}{GMM}{Gaussian Mixture Model}
\newacronym{gmr}{GMR}{Gaussian Mixture Regression}
\newacronym{dtw}{DTW}{Dynamic Time Warping}
\newacronym{pbd}{PBD}{Probgramming by Demonstration}
\newacronym{hic}{HIC}{Hybrid Impedance Control}
\newacronym{rl}{RL}{Reinforcement Learning}
\newacronym{rass}{RASS}{Robotically Assisted Surgical Systems}
\newacronym{mav}{MAV}{Mean Absolute Value}
\newacronym{avf}{AVF}{Adaptive Virtual Fixture}
\newacronym{tcp}{TCP}{Tool Center Point}
\newacronym{lmm}{LMM}{Linear Mixed Model}
\newacronym{iqr}{IQR}{Interquartile Range}
\newacronym{or}{OR}{Operating Room}
\newacronym{ltb}{LTB}{Lübeck Toolbox}
\newacronym{sus}{SUS}{System Usability Scale}
\begin{document}

\title{EMG-Based Adaptation of Anisotropic Virtual Fixtures \\for Robot-Assisted Surgical Resection and Dissection}
\author{Dario Onfiani$^{1}$, Michael Dyck$^{2}$, Luigi Biagiotti$^{1}$ and Julian Klodmann$^{2}$
\thanks{$^{1}$D. Onfiani and L. Biagiotti are with the Department of Engineering "Enzo Ferrari", University of Modena and Reggio Emilia, Italy (e-mail: dario.onfiani@unimore.it, luigi.biagiotti@unimore.it).}
\thanks{$^{2}$M. Dyck and J. Klodmann are with the Institute of Robotics and Mechatronics, German Aerospace Center (DLR), Munich, Germany (e-mail: michael.dyck@dlr.de, julian.klodmann@dlr.de).}
}
%--- BLOCCO AUTORI ORIGINALE (COMMENTATO PER DOUBLE-BLIND) ---
\author{Dario Onfiani$^{1}$, Michael Dyck$^{2}$, Luigi Biagiotti$^{1}$ and Julian Klodmann$^{2}$
\thanks{$^{1}$D. Onfiani and L. Biagiotti are with the Department of Engineering "Enzo Ferrari", University of Modena and Reggio Emilia, Italy (e-mail: dario.onfiani@unimore.it, luigi.biagiotti@unimore.it).}
\thanks{$^{2}$M. Dyck and J. Klodmann are with the Institute of Robotics and Mechatronics, German Aerospace Center (DLR), Munich, Germany (e-mail: michael.dyck@dlr.de, julian.klodmann@dlr.de).}
\thanks{Experiments approved by the DLR ethics committee (Ref. 7/26)} % <-- DA DECOMMENTARE NELLA CAMERA-READY
}

% % --- BLOCCO AUTORI ANONIMO PER LA PRIMA SOTTOMISSIONE ---
% \author{Anonymous Authors % Sostituisci con "Paper ID: XXXX" se richiesto dalle linee guida della conferenza/journal
% \thanks{[Ethical approval details omitted for double-blind review.]}

%}

\maketitle
\begin{abstract}
In this paper, we address the development of an adaptive assistance system for robot-assisted \textcolor{black}{laparoscopic} surgery, specifically for delicate \textcolor{black}{tasks} such as \acrlong{rd}. Even if \acrlong{vf}s offer significant advantages for guiding a surgeon's movements, conventional \acrlong{vf}s are often defined by fixed geometries, lacking the flexibility to adapt to the surgical workflow or the surgeon's immediate intent.
To address these limitations, we propose a novel framework for an adaptive and anisotropic virtual fixture. In addition, we introduce an intuitive control interface that modulates the fixture's geometry in real-time based on the surgeon's intent, inferred from \acrlong{emg} signals. This approach allows the surgeon to dynamically expand or disengage the constraint by contracting their forearm muscles, enabling seamless transitions between precise guided motion and free repositioning of the tool.
Experimental results from a \textcolor{black}{pilot} user study, based on a standardized surgical training task, demonstrate the effectiveness of the proposed method. The system showed significant improvements in task accuracy and movement consistency, alongside a reduction in perceived cognitive load, effort, and frustration.
\end{abstract} 

% \begin{IEEEkeywords}
% Article submission, IEEE, IEEEtran, journal, \LaTeX, paper, template, typesetting.
% \end{IEEEkeywords}

\section{Introduction}
% \input{Sections/1.1_NEW_Introduction2}
%\input{First_Submission_Sections/1_Introduction}
%last update 2026/02/05

%%%INTRODUCTION%%
\vspace{0 cm}
\acrfull{mis} has deeply transformed modern clinical practice by reducing patient trauma, accelerating recovery, and improving cosmetic outcomes~\cite{tonutti2017role,dagnino2024robot}. These advantages, however, come at the expense of increased technical demands: surgeons must operate with limited workspace, restricted haptic feedback, and indirect visualization, which substantially increases cognitive and physical workload, as consistently highlighted in surgical ergonomics studies~\cite{van2009optimal}.
Robotic assistance has been progressively introduced to mitigate these challenges, providing enhanced dexterity, motion scaling, and stereoscopic vision~\cite{klodmann2021introduction}. Beyond these well-established advantages, robotic platforms also open the possibility of integrating adaptive assistance mechanisms. In this broader framework, \acrfull{rass} are progressively evolving, combining ergonomic improvements with semi-autonomous support functions to enhance surgical precision, to reduce cognitive load, and enable more standardized outcomes across varying surgeon skill levels.
In this regard, \acrfull{rd} procedures represent a particularly challenging scenario. These tasks are characterized by delicate and often oscillatory ("plucking-like") tool motions, often performed under tissue tension and in close proximity to sensitive anatomical structures. The surgeon must not only maintain precise alignment along the intended dissection line but also frequently detach and re-approach the target region to adjust the entry angle or inspect the resection line. Crucially, safe surgical practice requires that surgeons must never cut blindly; thus, vertical motions of the surgical tool are frequently required to regain line of sight or limit penetration depth. This workflow reflects the known decomposition of complex surgical procedures into a ``grammar'' of fine-grained surgemes~\cite{van2021gesture}. This context highlights the need for guidance mechanisms that are both \emph{anisotropic}, providing selective freedom of motion along task-relevant directions, and \emph{dynamically adaptable}, allowing seamless adaptation of the constraint size, and transitions between guided and free motion~\cite{selvaggio2018passive}.
Virtual Fixtures (VF) have been introduced as a promising framework for facilitating such tasks, by modulating the operator’s motion through real-time constraints or potential fields enforced by the robot during shared-control~\cite{bowyer2013active,abbott2007haptic}.
While this strategy has demonstrated benefits in highly-structured applications, such as in orthopedic surgery, where guidance is applied relative to pre-operative models of rigid anatomy~\cite{davies1997active}, their direct application to laparoscopic \acrshort{rd} tasks on soft tissue remains limited. In particular, the deformable nature of the anatomy, the anatomical complexity, the variability of dissection strategies, and the need for frequent re-approaches requires guidance mechanisms that can be adapted online to both the surgical context and the surgeon’s intent.
\begin{figure}
    \centering
        \includegraphics[width=1\columnwidth]{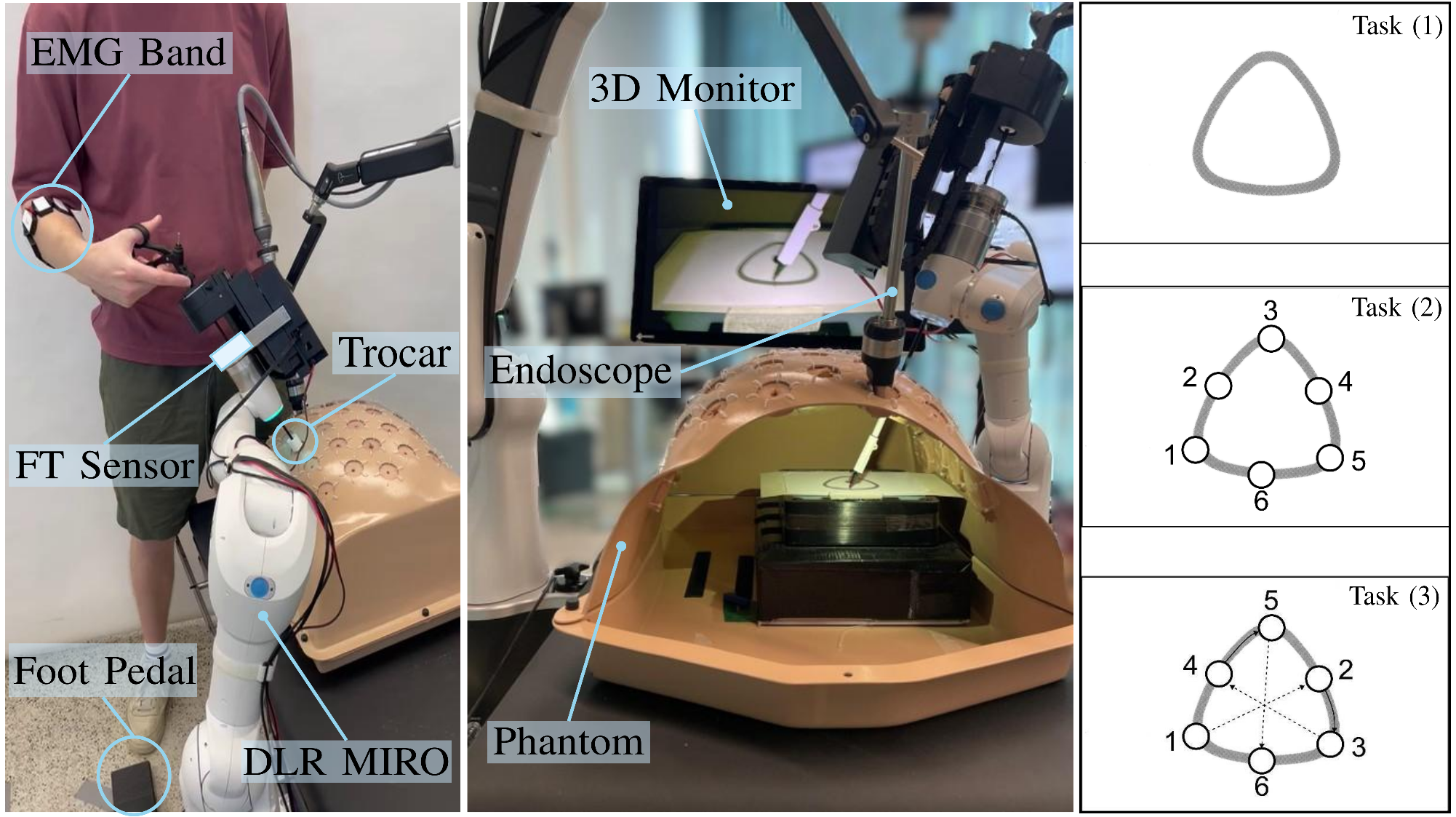}
        {\hspace{6mm}\small (a) \hspace{3.3cm}(b) \hspace{2.5cm}(c)}
        \caption{Experimental overview: (a) the robotic platform and user interface, (b) the custom drawing apparatus for the validation study, (c) standardized tasks mimicking surgical surgemes.
        }
    \label{fig:exp_setup_a}
\end{figure}
%
%%%RELATED WORKS%%
\subsection*{Related Work}
\acrshort{vf}s are categorized into two main classes: \textit{non-adaptive} and \textit{adaptive}~\cite{bowyer2013active}. 
Non-adaptive \acrshort{vf}s offer robust guidance but lack flexibility to handle changes in the environment or in the surgical workflow. They are typically defined pre-operatively using 3D anatomical models~\cite{li2020anatomical,feng2021virtual} or vector fields for static safety boundaries~\cite{marinho2018active}. Intra-operative approaches generate fixtures directly within the workflow, relying on motion models like minimum-jerk trajectories~\cite{lin2019virtual} or manual point marking~\cite{feng2021virtual}. Vision-based methods have also been proposed for dynamic collision avoidance~\cite{moccia2020vision}. However, the inherent rigidity of these methods limits their ability to cope with tissue deformation or discontinuous surgical plans.
Adaptive strategies overcome these limitations by modifying constraints online. \textit{Environment-driven} adaptation employs vision algorithms to track anatomical targets and update the path in real-time~\cite{moccia2019vision}, while \textit{performance-driven} paradigms dynamically adjust fixture parameters, such as the guidance radius, based on automated real-time skill evaluation~\cite{shahbazi2013dual}.
Finally, \textit{user-driven} methods place the surgeon in control, allowing not only geometry re-definition via \acrfull{kt}~\cite{selvaggio2018passive} but also the modulation of the \acrshort{vf} to match the operator's voluntary engagement.
A direct physiological measure of such engagement is provided by \acrfull{emg} signals. Beyond estimating operator stiffness for learning-based automation~\cite{arduini2024learning}, \acrshort{emg} has been applied to real-time shared control. Existing approaches typically utilize signal magnitude to continuously regulate the assistance gain~\cite{fan2021regulable} or employ classifiers to trigger discrete switches between control modes~\cite{patriarca2025emg}, allowing the user to modulate the robot's behavior.
%
%%%CONTRIBUTIONS%%%
\subsection*{Contibutions}
\begin{figure}[t]
    \centering
    \includegraphics[width =1\columnwidth]{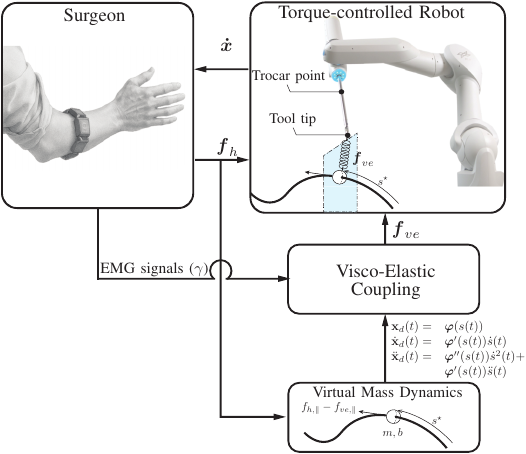} 
    \caption{Block-diagram of the proposed control architecture for robot-assisted \acrshort{rd}.}
    \label{fig:System_Overview}
\end{figure}
In this work, we propose a novel control architecture for adaptive virtual fixtures, depicted in Fig.~\ref{fig:System_Overview}. A reference path, defined via \acrshort{kt} by having the user guide the \acrfull{tcp}, serves as a constraint for a virtual mass, whose dynamics govern the robot's motion along the curve~\cite{Onfiani2025optimizing}. This path is utilized to generate anisotropic constraints, defining a guidance corridor with independent geometric bounds and stiffness properties along the normal and binormal directions~\cite{bettini2004vision}. These constraints restrict motion to the reference path with high precision, while preserving the necessary workspace for the oscillatory patterns inherent to \acrshort{rd} tasks.
To the best of our knowledge, this method advances the state of the art by introducing a flexible human-in-the-loop interface for direct, continuous \acrshort{vf} modulation. Specifically, the main contributions of this work are threefold: (i) the development of an anisotropic \acrlong{vf} based on a moving frame, enabling smooth execution of \acrshort{rd} tasks while restricting undesired motion; (ii) the integration of an \acrshort{emg}-based control mechanism that enables continuous adaptation of the constraint size based on the surgeon’s muscular co-contraction; and (iii) the design of a dynamic (dis-)engagement mechanism ensuring consistent initialization of virtual mass dynamics from any user-selected point along the path.
%last update 2026/02/12
\section{Adaptive Virtual Fixture Framework}
% \input{First_Submission_Sections/2_Adaptive_Virtual_Fixture_Framework} 
%last update 2026/02/05
\label{sec:Adaptive_virtual_fixtures}
\subsection{Reference Geometry and Frame Construction}
\begin{figure}[t]
    \centering
    \includegraphics[width=1\linewidth,height = 6cm]{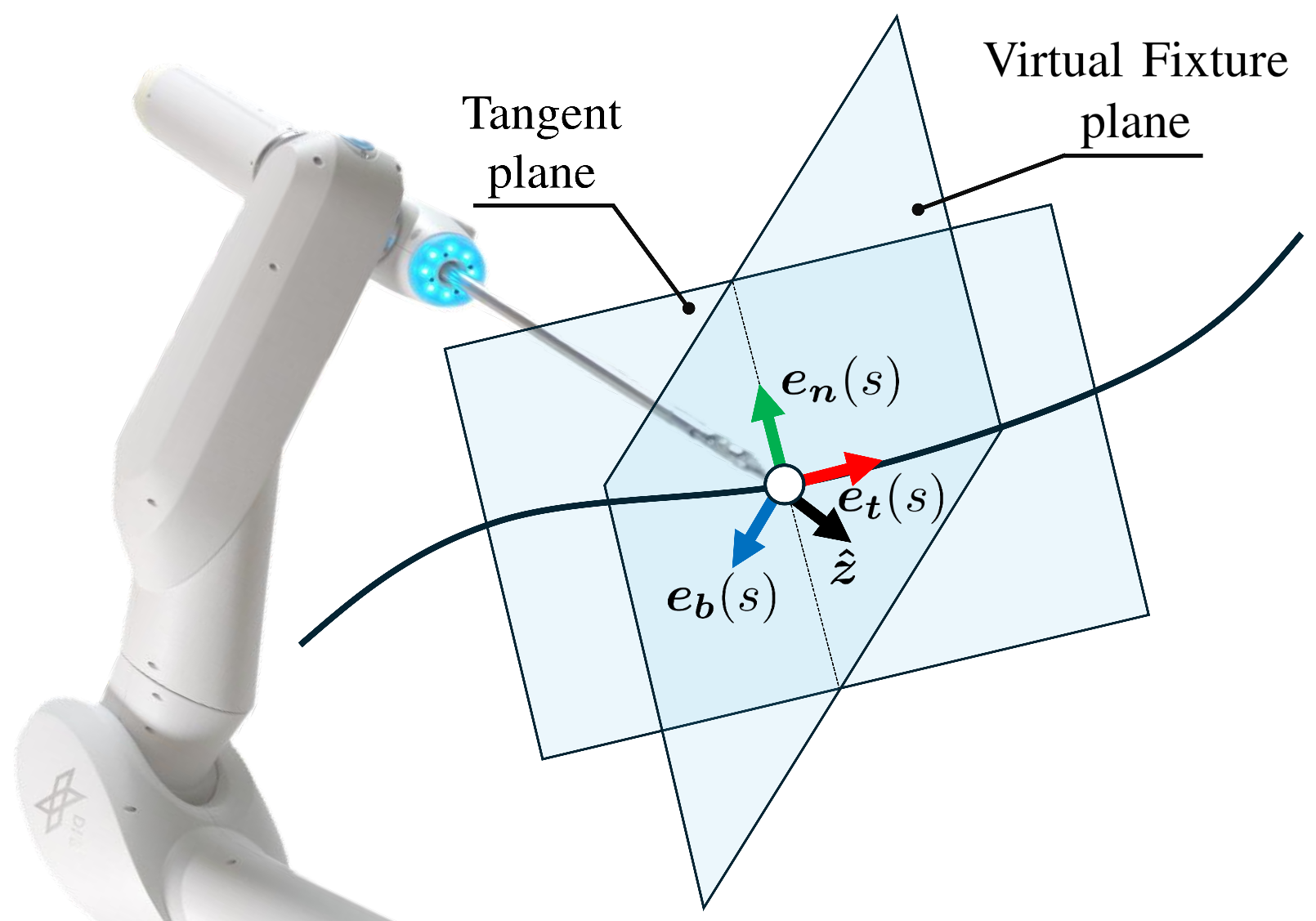}
    \caption{Reference path generation and frame construction.}
    \label{fig:Reference_Path_Generation_and_Frame_Construction}
\end{figure}
This section outlines the control architecture adopted to implement the proposed virtual fixture, following the design principle introduced in \cite{Onfiani2025optimizing}.  
As illustrated in Fig.~\ref{fig:System_Overview}, the force exchanged between the user and the robot is measured and applied to a virtual mass constrained to the desired Cartesian path
\begin{equation}
\boldsymbol{x}_d(t) = \boldsymbol{\varphi}(s(t)),
\end{equation}
parameterized by its arc length \( s \). A suitable viscoelastic coupling, detailed in Sec.~\ref{ssec:fvf},
\(\boldsymbol{f}_{ve}(\dot{\tilde{\boldsymbol{x}}}, \tilde{\boldsymbol{x}})\), with
\(\tilde{\boldsymbol{x}} = \boldsymbol{x} - \boldsymbol{x}_d\), where \(\boldsymbol{x} \in \mathbb{R}^3\) denotes the Cartesian position of the tool tip, is then defined between the virtual mass and the robot.
This additional force \(\boldsymbol{f}_{ve}\) can be integrated into any torque-based robot control scheme to enforce the virtual constraints. Conversely, \(\boldsymbol{f}_{ve}\) influences the dynamics of the virtual mass along the path according to
\begin{equation}
m \ddot{s}(t) + b \dot{s}(t)
= f_{h,\parallel} - f_{ve,\parallel}\bigl(\dot{\tilde{\boldsymbol{x}}}, \tilde{\boldsymbol{x}}\bigr).
\label{eq:MassDynamics}
\end{equation}
Obviously, only the tangential components of the user force
$
f_{h,\parallel}
= \bigl(\boldsymbol{{f}}_h\bigr)^{\!T} \boldsymbol{e}_t
$
and of the viscoelastic coupling \(f_{ve,\parallel}\) determine the progression \(s(t)\) of the virtual mass along the path, since the normal components are compensated by the reaction forces imposed by the constraint.
To generate the reference path, we utilize \acrshort{kt}~\cite{Onfiani2025optimizing}. The user guides the tool tip within the workspace, and the recorded positions are interpolated via cubic B-splines. Converting to arc-length parameterization yields a time-independent geometric description $\boldsymbol{\varphi}(s)$ of the task.
Subsequently, to enable an anisotropic fixture definition, a local orthonormal frame $\{\boldsymbol{e}_t, \boldsymbol{e}_n, \boldsymbol{e}_b\}$ attached to the virtual mass is constructed along this path (Fig.~\ref{fig:Reference_Path_Generation_and_Frame_Construction}). Unlike a standard Frenet frame, this basis explicitly accounts for the tool approach direction $\boldsymbol{\hat{z}}$ recorded during \acrshort{kt}.
The tangent vector $\boldsymbol{e}_{t}(s)$ aligns with the path derivative $d\boldsymbol{\varphi}/ds$. To facilitate task-specific anisotropy, the normal $\boldsymbol{e}_{n}(s)$ is defined by projecting $\boldsymbol{\hat{z}}$ onto the plane orthogonal to $\boldsymbol{e}_{t}$, while the binormal completes the right-handed basis as $\boldsymbol{e}_{b} = \boldsymbol{e}_{t} \times \boldsymbol{e}_{n}$. This construction allows defining independent constraints along the tool axis and its orthogonal plane.
\subsection{Anisotropic Virtual Fixture Formulation}
\label{ssec:fvf}
\begin{figure}[t]
    \centering
    \includegraphics[width=0.75\linewidth,height=6cm]{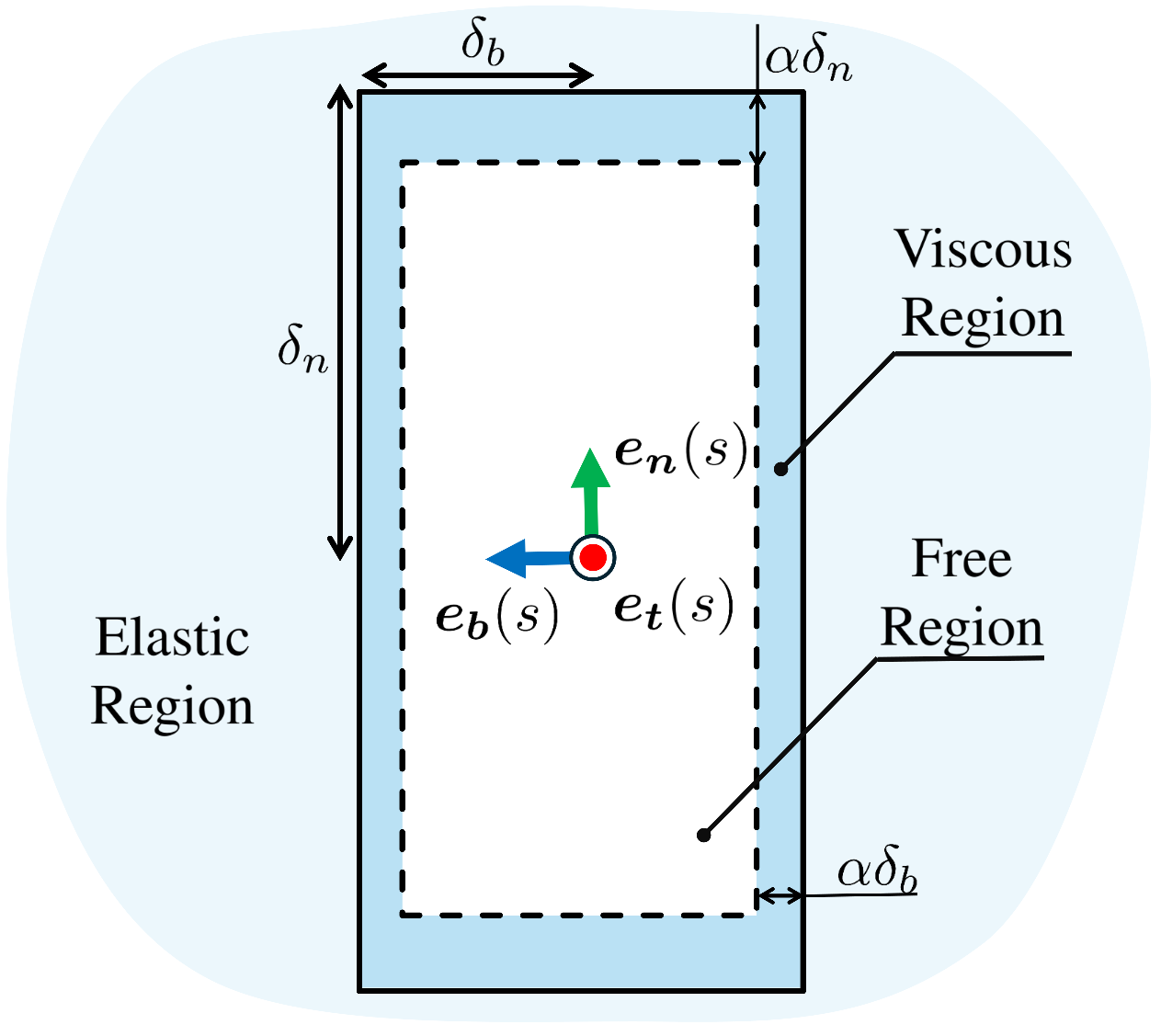}
    \caption{Schematic representation of the anisotropic constraint.}
    \label{fig:constraint_geomerty}
\end{figure}

The viscoelastic force is decomposed into tangential and orthogonal components:
\begin{equation}
\boldsymbol{f}_{ve}
= \boldsymbol{f}_{ve,\parallel} + \boldsymbol{f}_{ve,\perp}.
\label{eq:VEForce}
\end{equation}
The tangential component
\begin{equation}
\boldsymbol{f}_{ve,\parallel}
= \bigl( k_t \tilde{x}_t + d_t \dot{\tilde{x}}_t \bigr)\boldsymbol{e}_t
\end{equation}
enforces synchronization between the \acrshort{tcp} and the virtual mass.
The orthogonal component \(\boldsymbol{f}_{ve,\perp}\) constrains the surgeon motion within a virtual corridor spanned by the normal and binormal directions \(\boldsymbol{e}_n\) and \(\boldsymbol{e}_b\), respectively (Fig.~\ref{fig:constraint_geomerty}):
\begin{equation}
\boldsymbol{f}_{ve,\perp}
= f_{ve,n}\,\boldsymbol{e}_n + f_{ve,b}\,\boldsymbol{e}_b .
\label{eq:fperp}
\end{equation}
Each orthogonal component \(i \in \{n,b\}\) follows a nonlinear viscoelastic law
\begin{equation}
f_{ve,i}
= k_i\,\mathrm{dz}(|\tilde{x}_i|,\delta_i)
+ d_i(|\tilde{x}_i|,\delta_i)\,\dot{\tilde{x}}_i ,
\end{equation}
where the elastic term is defined through a \emph{dead-zone} function that specifies a free-motion region of width \(\delta_i\):
\begin{equation}
\mathrm{dz}(|\tilde{x}_i|,\delta_i)
=
\begin{cases}
0, & \text{if } |\tilde{x}_i| \le \delta_i, \\[4pt]
|\tilde{x}_i| - \delta_i, & \text{if } |\tilde{x}_i| > \delta_i .
\end{cases}
\label{eq:DeadFunction}
\end{equation}
To ensure smooth transitions at the boundary of the dead zone, the damping coefficient \(d_i\) is progressively activated within a transition band of width \(\alpha \delta_i\), with \(\alpha \in [0,1]\):
\begin{equation}
\footnotesize 
d_i(|\tilde{x}_i|,\delta_i)
=
\begin{cases}
D_i,
& \text{if } |\tilde{x}_i| > \delta_i, \\[8pt]
\displaystyle
\frac{|\tilde{x}_i| - (1-\alpha)\delta_i}{\alpha \delta_i}\, D_i,
& \text{if } (1-\alpha)\delta_i \le |\tilde{x}_i| \le \delta_i, \\[10pt]
0,
& \text{otherwise.}
\end{cases}
\end{equation}
This independent formulation along the normal and binormal directions results in a rectangular guidance corridor. While simple to implement, the sharp corners of this geometry may lead to abrupt variations in the constraint force. Alternative corridor shapes, such as elliptical or stadium-like geometries, could be adopted to obtain smoother boundaries.
%last update 2026/02/12
\section{EMG-Based Constraint Adaptation }
\label{sec:Constraint_Adaptation_via_EMG-Based Interaction}
This section describes how the user's intention, inferred from forearm muscle contractions, is used to dynamically adapt the virtual constraint.%
\subsection{Signal Processing}
\label{subsec:Signal_Processing}
To obtain a quantitative measure of the user's muscular contraction from the forearm, \acrshort{emg} signals were acquired using a MYO\textsuperscript{\textregistered} Armband (Thalmic Labs, Canada)~\cite{rawat2016evaluating} and processed in real-time. The raw multi-channel data, however is often noisy and requires a robust processing pipeline to be reliably interpreted. To obtain a robust scalar activation index $\gamma(t) \in [0,1]$, the raw \acrshort{emg} channels were processed via \acrlong{mav}, smoothed with a 0.5\,s moving average window to reject high-frequency noise, and finally normalized. 
\subsection{Constraint Adaptation via an \acrshort{emg}-Based Virtual Clutch}
\label{subsec:Constraint_Adaptation}
The dimensions of the virtual channel (Fig.~\ref{fig:constraint_geomerty}), denoted by $\delta_i$ ($i\in\{n,b\}$), are modulated in real-time by a virtual clutch logic driven by the activation index $\gamma(t)$. 
When the user wishes to disengage the constraint (e.g., to reposition the tool), they contract their forearm muscles. If the signal exceeds the threshold ($\gamma(t) \ge \bar\gamma$), the system activates a \textit{manual widening} mode: the constraint boundary $\delta_i(t)$ instantly expands to match the current deviation, $\delta_{i}(t) \leftarrow \max(|{\tilde{x}}_i(t)|, \bar\delta_{i})$, allowing resistance-free motion.
Conversely, when muscles are relaxed ($\gamma(t) < \bar\gamma$), the clutch triggers an \textit{automatic shrinking} mode. In this state, the boundary $\delta_i(t)$ tracks the user's motion only when they move towards the reference path ($d|\tilde{x}_i|/dt < 0$), progressively reducing $\delta_i$ until the nominal size $\bar\delta_i$ is restored. This unidirectional update logic prevents the user from experiencing sudden elastic forces upon re-engagement.
%last update 2026/02/05
\section{Experiments}
%\input{First_Submission_Sections/4_Experiments}
%last update 2026/02/05
\label{sec:Experiments}
To assess the effectiveness of the proposed \acrshort{emg}-based adaptive \acrlong{vf} in assisting complex \acrshort{rd} tasks, a pilot user study was conducted comparing the assistance method against free-motion execution.
The primary objective was to verify the efficacy of the system. It was hypothesized that the proposed solution would significantly improve \textnormal{task accuracy} and \textnormal{movement consistency} in path following and target reaching tasks compared to unassisted execution.
Furthermore, the adaptive assistance was expected to reduce \textnormal{cognitive workload} and \textnormal{perceived effort}, particularly in complex tasks requiring frequent transitions between continuous tracking and discrete positioning.
\subsubsection*{Sample}
The study involved \textnormal{seven participants} (four male and three female), with ages ranging from 24 to 29 years. All participants were right-handed. The group consisted of untrained users without a specific surgical background, chosen to evaluate the intuitiveness of the system for novice operators.
\subsubsection*{Apparatus}
The proposed VF framework was validated utilizing the setup depicted in Fig.~\ref{fig:exp_setup_a}. It includes a 7-DoF lightweight robotic arm, the DLR MIRO, part of the DLR MiroSurge system~\cite{seibold2019dlr}, a research platform for robotic assistance in laparoscopic surgery. The DLR MIRO was equipped with a DLR telescopic linear axis~\cite{alex_dlr}, which serves as the drive unit for inserting and extracting conventional laparoscopic tools, typically including standard instruments such as grasping forceps (e.g., KARL STORZ CLICKline\footnote{KARL STORZ SE \& Co. KG, Tuttlingen, Germany\label{fn:KARLSTORZ}}). Both systems are torque-controlled at the joint level~\cite{le2007mimo,le2008friction}; therefore, they form a redundant robotic system with a total of 8~DoF, which can be modeled according to the standard Lagrangian formulation as
\begin{equation}
\boldsymbol{M}(\boldsymbol{q})\,\ddot{\boldsymbol{q}}
+
\boldsymbol{C}(\boldsymbol{q},\dot{\boldsymbol{q}})\,\dot{\boldsymbol{q}}
+
\boldsymbol{g}(\boldsymbol{q})
=
\boldsymbol{\tau}
+
\boldsymbol{\tau}_{ext},
\nonumber
\end{equation}
where \(\boldsymbol{q} \in \mathbb{R}^8\) denotes the vector of joint coordinates, \(\boldsymbol{M}(\boldsymbol{q})\) is the inertia matrix, \(\boldsymbol{C}(\boldsymbol{q},\dot{\boldsymbol{q}})\) collects Coriolis and centrifugal terms, \(\boldsymbol{g}(\boldsymbol{q})\) represents the gravity contribution, \(\boldsymbol{\tau}\) is the vector of joint torques, and \(\boldsymbol{\tau}_{ext}\) accounts for external torques induced by interaction forces with both environment and user.
A hierarchical impedance control structure with three levels, inspired by \cite{dietrich2019hierarchical
 }, has been implemented, commanding the desired joint torques:
\begin{equation}
    \boldsymbol{\tau}^{\star} =
    \boldsymbol{g}(\boldsymbol{q})
    + \boldsymbol{\tau}_{\tiny tp}
    + \boldsymbol{N}_{\!\tiny tp}\boldsymbol{\tau}_{\tiny tcp}
    + \boldsymbol{N}_{\!\tiny tp}\boldsymbol{N}_{\!\tiny tcp}\boldsymbol{\tau}_{\tiny j_3} \label{eq:HierarchicalControl}
\end{equation}
where
\begin{align}
    \boldsymbol{\tau}_{\tiny tp}
    &=
    \boldsymbol{J}_{\tiny tp}^{\top}
    \left(
        \boldsymbol{K}_{p}\,\accentset{\sim}{\boldsymbol{r}}_{\tiny tp}
        + \boldsymbol{K}_{d}\,\dot{\accentset{\sim}{\boldsymbol{r}}}_{\tiny tp}
    \right), \label{eq:T_tp} \\
    \boldsymbol{\tau}_{\tiny tcp}
    &=
    \boldsymbol{J}_{\tiny tcp}^{\top}\,\boldsymbol{f}_{ve}. \label{eq:T_tcp}
\end{align}
Here, $\accentset{\sim}{\boldsymbol{r}}_{\tiny tp}
= \boldsymbol{r}_{\tiny tp} - \boldsymbol{r}_{\tiny tp,d}$, where
$\boldsymbol{r}_{\tiny tp}
= [\,\boldsymbol{x}_{\tiny tp}^{\top}, \varphi_{\tiny tp}\,]^{\top}
= \boldsymbol{f}_{\tiny tp}(\boldsymbol{q})$
denotes the coordinates of the trocar point, including the position $\boldsymbol{x}_{\tiny tp}$ and the rotation angle $\varphi_{\tiny tp}$ about the tool $z$-axis\footnote{The rotary motion
about the tool shaft is done manually as the tool is attached
to the telescopic linear axis via a floating bearing.}.
The task Jacobian is computed as
\[
\boldsymbol{J}_{\tiny tp}(\boldsymbol{q})
= \frac{\partial \boldsymbol{f}_{\tiny tp}(\boldsymbol{q})}{\partial \boldsymbol{q}},
\]
and $\boldsymbol{K}_{p} = \mbox{diag}\{ k_{t}, k_{t}, k_{t}, k_{r}\} $, with the translational stiffness $ k_{t} = 2000 \frac{N}{m}$ and rotational stiffness $ k_{r} = 50 \frac{Nm}{rad}$. The damping matrix $\boldsymbol{K}_{d} $ is obtained by the double-diagonalisation design based on \cite{1242165} with a damping ratio of $\xi = 0.7$.  
Matrix $\boldsymbol{J}_{\tiny tcp}(\boldsymbol{q})$ is the Jacobian related to the tool center point position $\boldsymbol{x}$. Finally,  the torque null-space projectors $\boldsymbol{N}_{\!\tiny tp}$ and $\boldsymbol{N}_{\!\tiny tcp}$ are computed using the dynamically consistent formulation. 
For a generic task~$*$ with Jacobian $\boldsymbol{J}_{*}(\boldsymbol{q})$, the dynamically consistent pseudoinverse is defined as
\begin{equation}
    \boldsymbol{J}_{*}^{\#}(\boldsymbol{q})
    =
    \boldsymbol{M}^{-1}(\boldsymbol{q})\,
    \boldsymbol{J}_{*}^{\top}
    \left(
        \boldsymbol{J}_{*}\boldsymbol{M}^{-1}(\boldsymbol{q})\boldsymbol{J}_{*}^{\top}
    \right)^{-1}.
\end{equation}
The corresponding dynamically consistent null-space projector is given by
\begin{equation}
    \boldsymbol{N}_{\!*}
    =
    \boldsymbol{I}
    -
    \boldsymbol{J}_{*}^{\top}\boldsymbol{J}_{*}^{\#\top}.
\end{equation}
In conclusion, apart from gravity compensation, the first priority level in~\eqref{eq:T_tp} enforces the fulcrum constraint at the trocar point (TP).
The second priority level, given by \eqref{eq:T_tcp}, provides the interface for applying the viscoelastic force $\boldsymbol{f}_{ve}$, enforcing the virtual fixture at the tip of the instrument.
Finally, the redundancy of the MIRO robot is resolved through a joint impedance control $\boldsymbol{\tau}_{\tiny j_3} $ applied to joint~3.
With this impedance-based formulation, the robot behaves as a transparent interface:
in the absence of VF forces, the user perceives the instrument behavior as that of a conventional free-floating laparoscopic tool constrained only by the trocar, without significant resistance from the robotic structure.
\\
A phantom of an insufflated abdomen (LapTrainer\footnote{Insufflated Abdomen with Ports (\#2213), The Chamberlain Group, Great Barrington, USA}) was positioned on an \acrfull{or} table. This LapTrainer had various openings for the placement of trocars. To mantain a standardized endoscope position for every participant, a SOLOASSIST\textsuperscript{II}\footnote{AKTORmed GmbH, Neutraubling, Germany} robotic camera control was used to held the stereoscopic endoscope KARL STORZ TIPCAM\textregistered{}1 Rubina\textregistered{} 4K/3D\footref{fn:KARLSTORZ} used for the study. The endoscopic video stream was displayed on a standard stereoscopic display positioned on the \acrshort{or} table.
For the specific validation purposes of this study, instead of using an actual surgical tool, a custom tool was designed by attaching a permanent marker to the instrument tip. This allowed converting a surgical task into a path-following drawing task on a flat surface (Fig.~\ref{fig:exp_setup_a}). \textcolor{black}{Additionally, a foot pedal was integrated into the setup to allow the user to voluntarily enable robot control and initiate task execution safely.}
Finally, to estimate user intention and modulate the constraint, a Myo Armband was attached to the user's dominant arm.
\subsubsection*{Experimental Task and Design}
Guidance Virtual Fixtures hold significant potential for assisting in complex R\&D tasks.
    \textcolor{black}{To validate the proposed control architecture in a representative yet standardized scenario, we selected a benchmark that demands both high continuous path-following precision and the ability to handle sharp directional changes. Therefore,} the experimental task was adapted from the cutting exercises of the Fundamentals of Laparoscopic Surgery and the \textit{'Triangle Cut'} exercise of the \acrfull{ltb} curriculum~\cite{LubeckToolboxWebsite, thomaschewski2020efficacy}, as well as recent validated robotic training curricula such as \textit{RoSTraC~\cite{thomaschewski2024conception}}.
The study followed a \textnormal{within-subjects design}~\cite{budiu2018between}, where each participant performed the tasks under both experimental conditions.
\textcolor{black}{To avoid carry-over effects from the guidance system to the unassisted performance, the Free Motion condition was always presented first to establish a baseline. To mitigate potential learning effects resulting from this fixed order, a dedicated familiarization phase was conducted prior to data collection, allowing participants to practice until they were comfortable with the task dynamics.}
To ensure the simplified experimental design remained representative of the real surgical procedure, we identified three fundamental motor elements (or \textit{surgemes}\cite{van2021gesture}) characteristic of dissection: \textit{continuous cutting}, \textit{discrete ``plucking-like'' motions}, and the \textit{repositioning} of the tool along the resection line. Consequently, three distinct experimental tasks were defined (see Fig.~\ref{fig:exp_setup_a})%(see Fig.~\ref{fig:task_patterns}):
\begin{itemize}
    \item \textbf{Task 1} requires the user to continuously trace the reference curve. %(Fig.~\ref{fig:task1}).
    \textcolor{black}{Participants were allowed to choose their preferred travel direction to maximize biomechanical comfort. Since the tracking error and the virtual fixture assistance are independent of the motion direction, this degree of freedom does not introduce a confounding variable.}

    \item \textbf{Task 2} requires the user to mark the center of a set of circular targets distributed clockwise along the path. %(Fig.~\ref{fig:task2}).
    In this task, the user operates utilizing the predefined anisotropic constraint, without the need to adapt its size.

    \item \textbf{Task 3} combines elements of Task 1 and Task 2, %(Fig.~\ref{fig:task3}),
    requiring the user to alternately mark the target centers and connect them by drawing continuous lines. This mimics real surgical scenarios where tools are repositioned to adjust tissue retraction or inspection angles. To execute these transitions, the user must utilize the EMG-based virtual clutch to dynamically (dis-)engage the constraint.
\end{itemize}
%
% \begin{figure}[t]
%     \centering
%     \begin{subfigure}{0.3\linewidth}
%         \centering
%         \includegraphics[width=\linewidth]{img/img_test_setup_eps/task1_figure.eps}
%         \caption{}
%         \label{fig:task1}
%     \end{subfigure}
%     \hfill
%     \begin{subfigure}{0.3\linewidth}
%         \centering
%         \includegraphics[width=\linewidth]{img/img_test_setup_eps/task2_figure.eps}
%         \caption{}
%         \label{fig:task2}
%     \end{subfigure}
%     \hfill
%     \begin{subfigure}{0.3\linewidth}
%         \centering
%         \includegraphics[width=\linewidth]{img/img_test_setup_eps/task3_figure.eps}
%         \caption{}
%         \label{fig:task3}
%     \end{subfigure}
%     \caption{Reference patterns for the three experimental tasks proposed to participants. From left to right: (a) Task 1 (b) Task 2 (c) Task 3.}
%     \label{fig:task_patterns}
% \end{figure}
%
\subsubsection*{Procedure, Objective and Subjective Measures}
\textcolor{black}{A standardized procedure was followed for all participants. Prior to each recording block, participants underwent a dedicated familiarization phase lasting approximately 30 to 60 seconds. During this time, they practiced the specific task dynamics—either free-hand drawing or engaging the EMG-clutch—until they reported feeling comfortable with the system.} Subsequently, the actual performance was recorded. In this study, a \textnormal{single trial} is defined as the complete execution of one specific task under one condition.
To quantify the system's performance, kinematic and dynamic data were logged during the execution of each task. The primary \textit{objective metrics} defined for the evaluation include:
\begin{itemize}
    \item \textit{Tracking Error:} defined as the deviation between the \acrshort{tcp} position and the reference path when touching the paper. Specifically, for Task 2 and 3, the error component binormal to the path ($\boldsymbol{\tilde{x}_b}$) is analyzed to assess targeting precision.
    \item \textit{Interaction Forces:} the forces $\boldsymbol{f_h}$ exerted by the user are recorded to evaluate physical effort and the transparency of the assistance.
\end{itemize}
For \textit{subjective evaluation}, participants were asked to complete the NASA Task Load Index (NASA-TLX) questionnaire~\cite{hart1988development} at the end of each trial, to assess their perceived workload. This ensure that the workload evaluation reflects the immediate perception of each specific task, thereby minimizing recall bias. The tasks were presented to all participants in the same fixed sequence.
Once all tasks were completed, each user was asked to fill in the \acrfull{sus} questionnaire~\cite{brooke1996sus}. This combination of objective and subjective evaluations allowed for a comprehensive analysis of both system performance and user experience.
{\color{black}
\subsubsection*{Statistical Analysis}
Statistical methods were selected based on data distribution, verified via \textnormal{Shapiro-Wilk tests}.
For \textnormal{objective metrics}, due to the limited sample size ($N=7$) and deviations from normality observed in specific experimental conditions and paired differences, a non-parametric \textnormal{Wilcoxon signed-rank test} was consistently employed for all pairwise comparisons.
In contrast, \textnormal{subjective workload ratings} (NASA-TLX) were analyzed using \acrfull{lmm} to accommodate the $2 \times 3$ factorial design, after confirming that model residuals followed a normal distribution ($p > 0.05$). The model included \textit{Condition} and \textit{Task} (and their interaction) as fixed effects, while \textit{Subject} was modeled as a random intercept. This structure accounts for subject-specific baselines and rating variability, ensuring robust estimates despite the small sample size~\cite{baayen2008mixed}. Additionally, \textnormal{SUS scores} exhibited a normal distribution ($p=0.32$), justifying the use of descriptive mean statistics.
Significance levels were determined using Type III F-tests for \acrshort{lmm}s and standard p-values for Wilcoxon tests. All reported p-values are \textnormal{two-tailed} with a significance threshold of $\alpha < 0.05$.}

%last update 2026/02/12
\section{Results and Discussions}
% \input{First_Submission_Sections/5_Results}
%last update 2026/02/05
\label{sec:Result_and_Discussions}
Given the distinct nature of the experimental tasks, results are presented and immediately discussed within their specific context to facilitate interpretation.
\subsection{Objective Evaluation}
\label{subsec:objective_eval}
This section presents the performance metrics evaluated during the execution of the tasks described in Section~\ref{sec:Experiments}.
\subsubsection{Task 1 - Continuous Tracking}
\begin{figure}[t]
    \centering
    \includegraphics[width=\columnwidth]{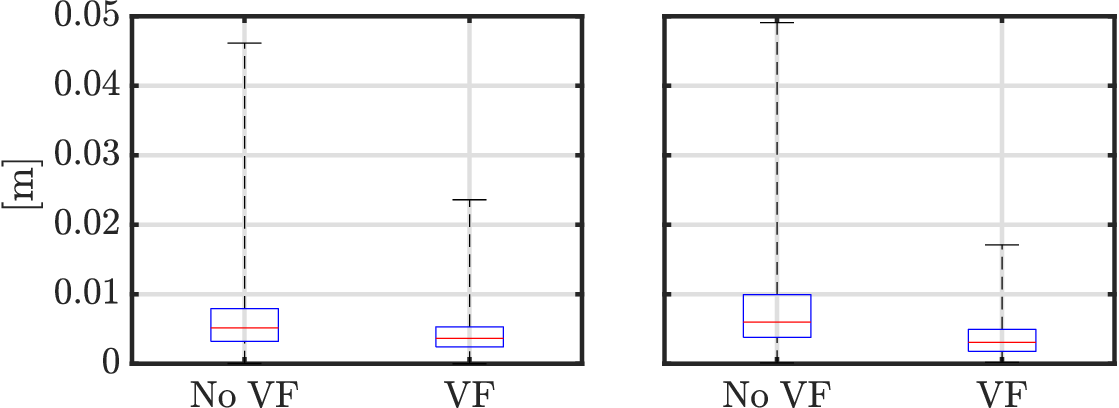}\\
    {\hspace{8mm}\small (a) \hspace{4cm}(b)}
    \caption{Deviation with respect to the reference path, obtained during the execution of (a) Task 1 and (b) Task 3.}
    \label{fig:position_error_task1_and_task3}
\end{figure}

We assessed tracking performance by computing the deviation between the actual \acrshort{tcp} position $\boldsymbol{x}$ and reference trajectory $\boldsymbol{x_d}$ in the $xy$-plane. Since $\boldsymbol{x_d}$ was generated via \acrlong{kt} using the robot kinematics, this metric directly quantifies the user's adherence to the intended path. 
Figure~\ref{fig:position_error_task1_and_task3}.a displays the distribution of the pointwise position error for Task 1. \textcolor{black}{As illustrated, the application of the \acrshort{vf} led to a significant reduction in tracking error ($p < 0.05$), with an average decrease of 27.7\% compared to the free motion condition.}%($W = 26.0, p = 0.047$)
Since Task 3 combines elements derived from both Task 1 and Task 2, we evaluated the same positional error metric for the path-following segments of Task 3 (Fig.~\ref{fig:position_error_task1_and_task3}.b) to assess the effectiveness of the \acrshort{vf} in this more complex scenario.
Notably, the reduction in position error achieved through the use of the \acrlong{vf} is even more pronounced in Task 3 than in Task 1. \textcolor{black}{This improvement was found to be statistically significant, yielding the maximum possible test statistic for this sample size ($p < 0.05$).} On average, the constrained condition resulted in an improvement of 50.3\% compared to the free motion condition.%($W = 28.0, p = 0.016$)
This enhanced performance is attributable to the hybrid nature of Task 3. Unlike Task 1, which allows the user to maintain a continuous motion state and benefit from short-term adaptation to the path dynamics, Task 3 imposes frequent \textit{task-switching}. These interruptions prevent the user from settling into a steady tracking rhythm, making the stabilization provided by the \acrshort{vf} crucial to compensate for the cognitive and motor overhead during transitions.\\
\subsubsection{Task 2 - Descrete Target Reaching}
\begin{figure}[t]
    \centering
    \includegraphics[width=1\columnwidth]{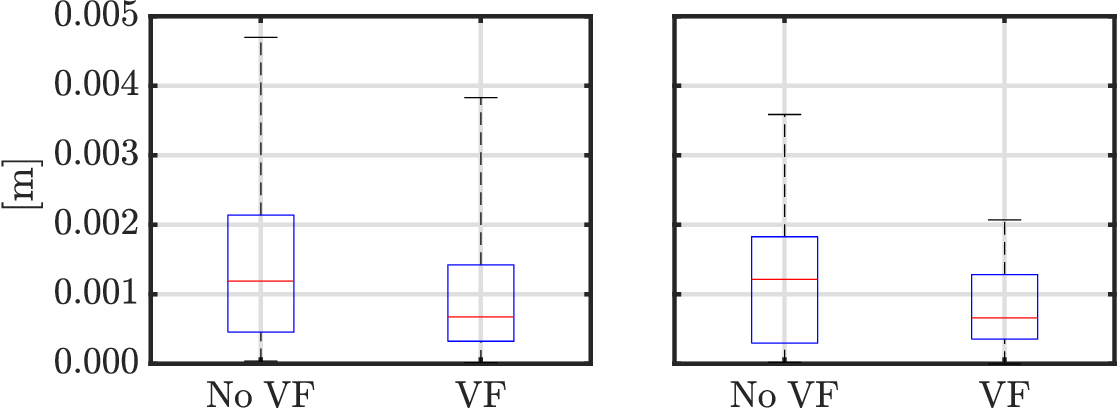}
    {\hspace{8mm}\small (a) \hspace{4cm}(b)}
    \caption{{Boxplots of the binormal component of the position error  \(\tilde{x}_b\) in (a) Task~2 and (b) Task~3.}}
    \label{fig:center_position_error_N_task2_and_task3}
\end{figure}

To assess the precision in reaching the predefined targets, the analyzed metric is defined as the position error $\tilde{x}_b$ from the target center along $\boldsymbol{e_{b}}$.
As shown in Fig.~\ref{fig:center_position_error_N_task2_and_task3}a, the \acrshort{vf} condition resulted in a notable performance improvement in Task 2, producing a reduction of 30.1\% in the binormal error component. % and a global mean error reduction of 23.3\%.
\textcolor{black}{Although the statistical analysis did not reach the significance threshold ($p = 0.109$), the substantial effect size indicates a clear trend toward improved precision.}
Specifically, for $\tilde{x}_b$, the \acrfull{iqr} decreased from 0.0017\,m to 0.0011\,m, marking a 34.6\% reduction. This result indicates a substantial improvement in execution consistency and repeatability when using the assistance. \textcolor{black}{This reduction in variability suggests that the lack of statistical significance in the mean comparison may be attributed to the limited sample size rather than a lack of system efficacy.}
These results confirm that the \acrshort{vf} not only mitigates the error magnitude but also improves the consistency of the user's behavior across trials, especially in the direction orthogonal to the reference path where the constraint is most effective.
Results for the targeting subtasks within Task 3,collected in Fig.~\ref{fig:center_position_error_N_task2_and_task3}b, mirror the trends observed in Task 2. The \acrshort{vf} condition yielded a 29.8\% reduction in mean error and a 39.4\% reduction in \acrshort{iqr}.
\textcolor{black}{Crucially, in this scenario involving task-switching, statistical analysis confirmed that the improvement in accuracy is significant ($p < 0.05$). This reinforces the finding that the \acrshort{vf} provides robust assistance particularly when the cognitive and motor demand increases.}%($W = 26.0, p = 0.047$)
\subsubsection{Task 3 - Constraint Size Adaptation}
\begin{figure}[t]
    \centering
    % --- Prima sottofigura (a sinistra) ---
    % Nota: la label va dentro le parentesi quadre []
    \subfloat[\label{fig:task3_path}]{
        \includegraphics[width=0.48\columnwidth,height=4.0cm]{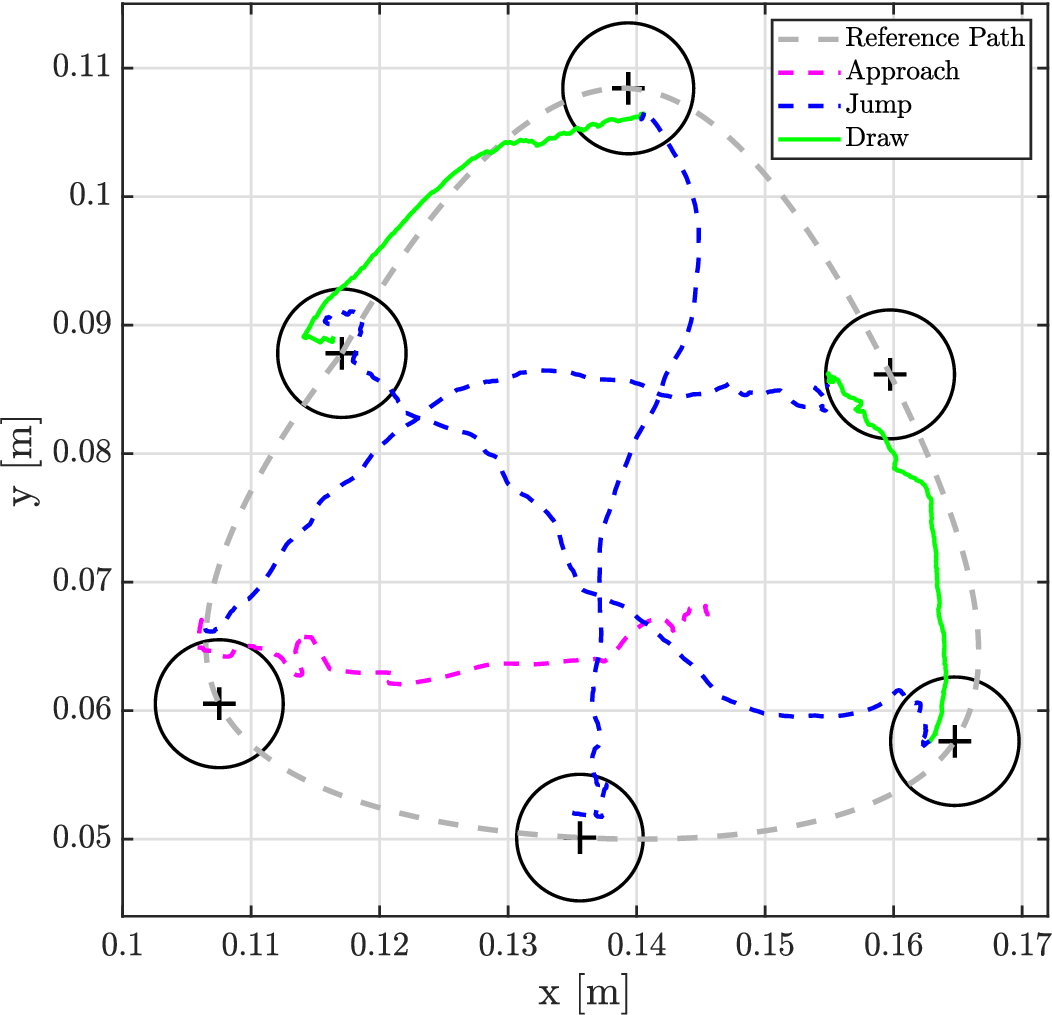}
    }
    % --- Seconda sottofigura (a destra) ---
    \subfloat[\label{fig:task3_emg}]{
        \includegraphics[width=0.48\columnwidth,height=4.0cm]{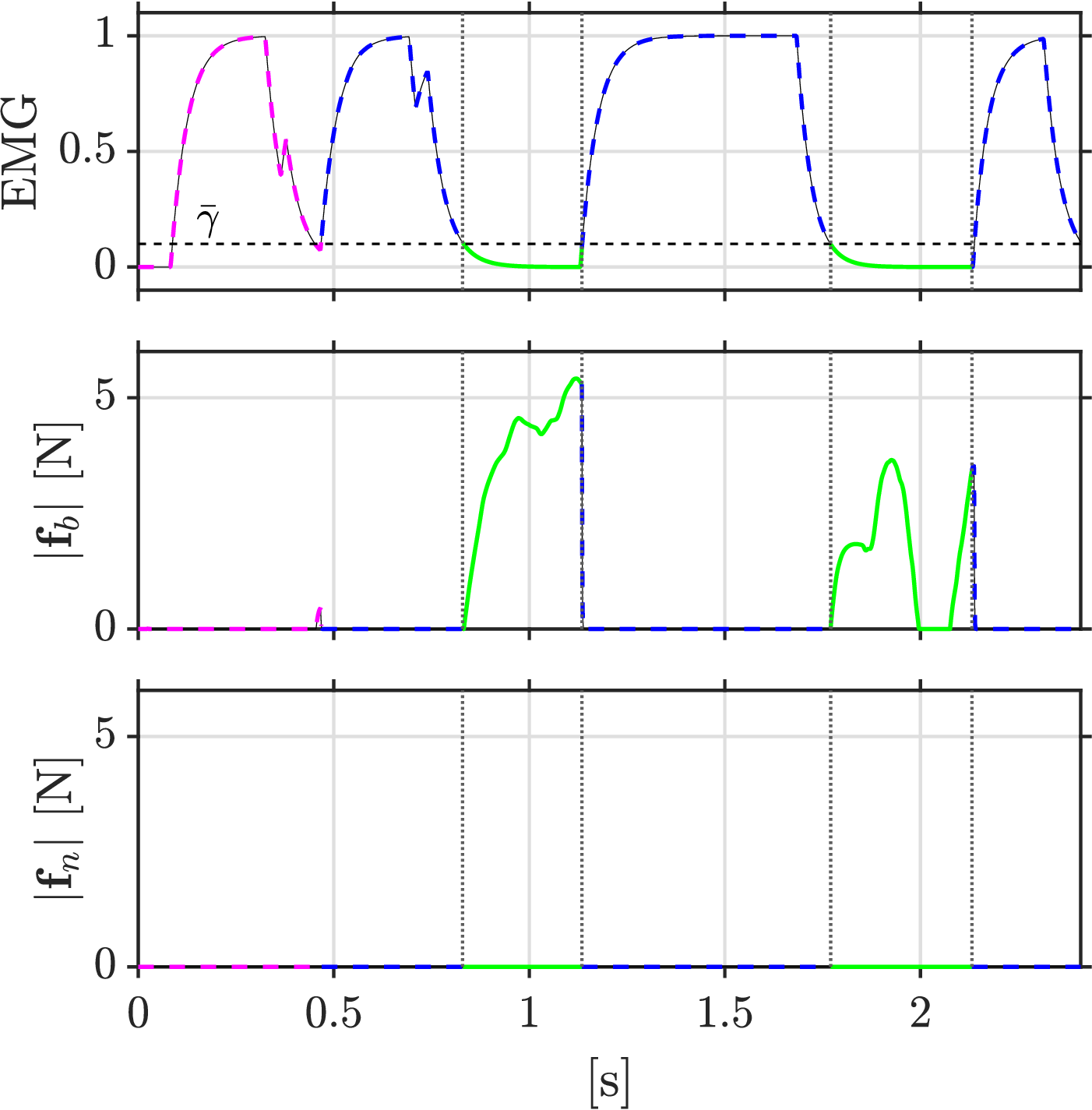}
    }
    
    \caption{Analysis of Task 3 execution. (a) Trajectory followed by the \acrshort{tcp}, highlighting transitions between constrained and unconstrained segments. (b) \acrshort{vf} normal and binormal forces vs. the user's EMG signal and system activation.}
    \label{fig:task3_analysis}
\end{figure}

We evaluated the effectiveness of the EMG-based adaptation mechanism. In this task, users were required to temporarily \textit{disengage} the constraint to transition between targets and \textit{re-engage} it for the continuous tracing segments.
The temporal alignment of the EMG activation peaks and the corresponding drop in constraint force (Fig.~\ref{fig:task3_analysis}.b) demonstrates the correct functionality of the proposed adaptation method (Sec.~\ref{subsec:Constraint_Adaptation}).
Specifically, whenever the EMG signal $\gamma(t)$ exceeds the threshold $\bar{\gamma}$, the virtual clutch triggers the \textit{disengaged state}, forcing the constraint force to null and allowing free repositioning. Conversely, as the signal drops, the clutch \textit{re-engages}. Notably, during the subsequent approach towards the reference path, the user perceives no resistive force despite the constraint being active, ensuring a smooth re-entry. During the tracking phase, the force is primarily exerted along $\boldsymbol{e_b}$, consistent with the intended guidance geometry.
To assess the physical burden of this interaction, we compared the average user-applied forces along the normal (\( \boldsymbol{f_{h,n}} \)) and binormal (\( \boldsymbol{f_{h,b}} \)) directions across the entire population (Table~\ref{tab:avg_forces}). Here, \( \boldsymbol{f_h} \) denotes the vector of interaction forces estimated from joint torque measurements~\cite{haddadin2008collision}, expressed in the workspace reference frame.
Statistical analysis revealed two distinct behaviors consistent with the anisotropic design of the virtual fixture. 
For the \textit{normal} component, corresponding to the direction of allowed motion, no significant difference was found ($p>0.05$), confirming that the assistance remains perfectly transparent where the user is intended to move freely. %($W = 15.0, p = 0.938$)
Conversely, for the \textit{binormal} component, which aligns with the stricter constraint boundaries, a statistically significant increase in interaction force was observed ($p < 0.05$), reflecting the active corrective action of the fixture against deviations.%($W = 0.0, p = 0.016$)
\textcolor{black}{This indicates that users actively utilized the constraint boundary for stabilization orthogonal to the path. However, as the absolute force magnitude remains below 1\,N, this interaction reflects helpful guidance rather than resistive burden.}
Overall, this lack of significant force increase combined with the low absolute values indicates that the assistance is transparent: the user accepts the guidance during tracking and, crucially, the virtual clutch disengages sufficiently fast during transitions to prevent the user from having to overcome resistive forces to break free.
\subsection{Subjective Evaluation}
\label{subsec:subjective_evaluation}
This section reports the users' perception of the system. We analyzed the workload for each specific task using the NASA-TLX questionnaire, specifically focusing on \textit{mental demand}, \textit{physical demand}, \textit{effort}, and \textit{frustration}. These dimensions were selected as they directly reflect the cognitive and physical challenges, as well as the overall perceived difficulty and stress associated with the task execution under both experimental conditions. Finally, the overall system usability was assessed via the \acrshort{sus} questionnaire.
\begin{table}[t]
\centering
\caption{\textsc{Average user-applied forces along constraint directions}}
\begin{tabular}{lcc}
\hline
\textbf{Condition} & \(\lvert f_{h,n} \rvert\) [N] & \(\lvert f_{h,b} \rvert\) [N] \\
\hline
No VF    & 0.817 & 0.696 \\
VF       & 0.868 & 0.848 \\
\hline
\end{tabular}
\label{tab:avg_forces}
\end{table}
\subsubsection{NASA-TLX}
\begin{figure}[t]
    \centering
    % --- Prima Figura (Radar Plot) ---
    \subfloat[\label{fig:nasa_radar}]{
        \includegraphics[width=0.48\columnwidth,height=3cm]{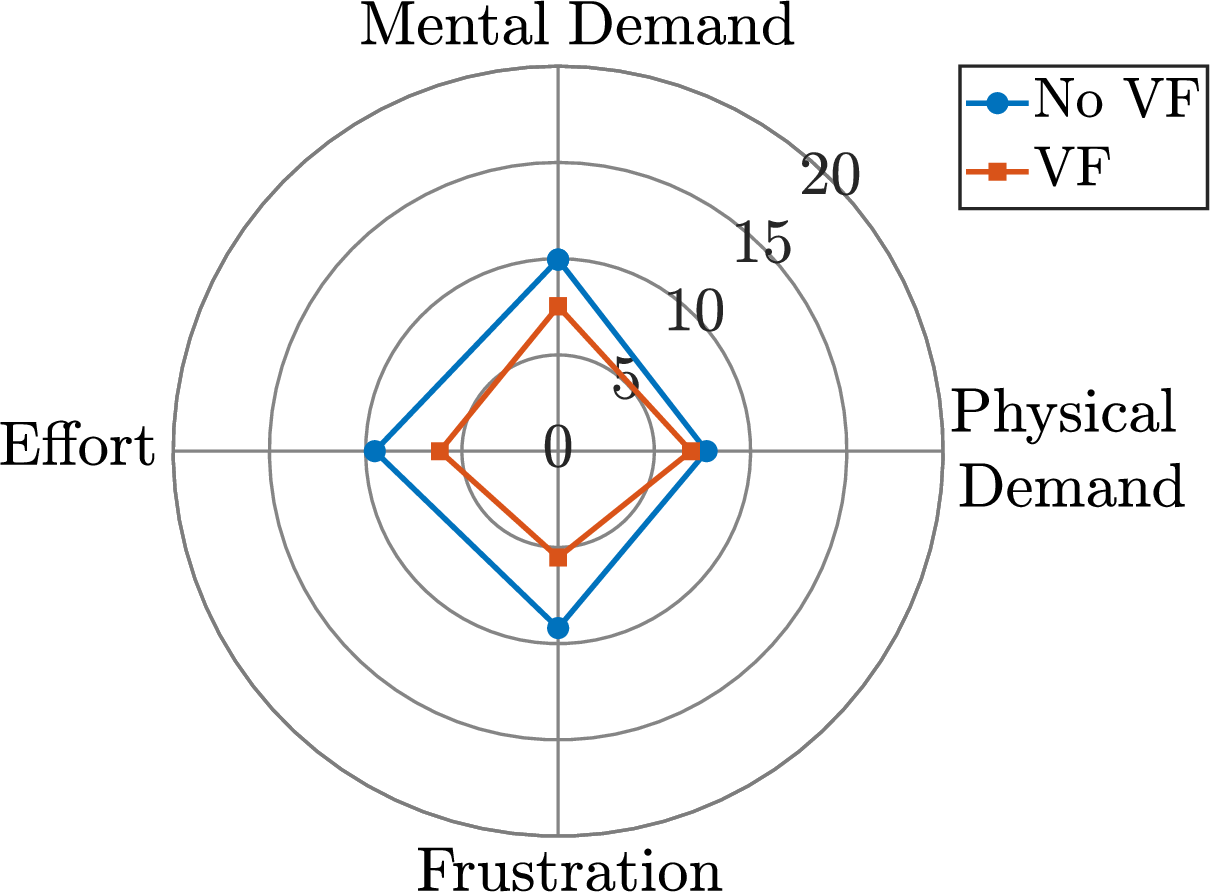}
    }
    % --- Seconda Figura (LMM Plot) ---
    \subfloat[\label{fig:estimated_means}]{
        \includegraphics[width=0.48\columnwidth,height=3cm]{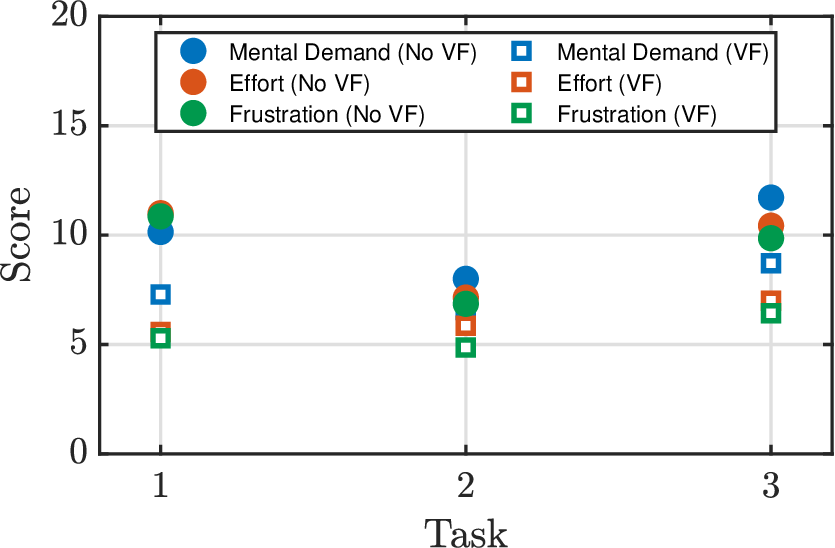}
    }
    
    \caption{(a) Subjective evaluation scores from NASA-TLX. (b) Estimated marginal means of NASA-TLX scores \textcolor{black}{for the statistically significant dimensions ($p < 0.05$).}}
    \label{fig:subjective_eval_combined}
\end{figure}

{\color{black}
Fig.~\ref{fig:estimated_means} summarizes the subjective workload ratings reported by participants.
The \acrshort{lmm} analysis revealed a significant main effect of the experimental \textit{Condition} across the key cognitive metrics. Specifically, the use of Virtual Fixtures significantly reduced \textit{Mental Demand} ($F(1, 36) = 6.23, p < 0.05$), \textit{Effort} ($F(1, 36) = 16.15, p < 0.05$), and \textit{Frustration} ($F(1, 36) = 20.47, p < 0.05$).
%\textit{Mental Demand} ($F(1, 36) = 6.23, p = 0.017$)
%\textit{Effort} ($F(1, 36) = 16.15, p < 0.001$), and \textit{Frustration} ($F(1, 36) = 20.47, p < 0.001$).
%
Conversely, no significant main effect was found for \textit{Physical Demand} ($F(1, 36) = 3.32, p > 0.05$). This subjective perception aligns with the objective interaction force data reported in Table~\ref{tab:avg_forces}, suggesting that the active guidance did not result in a statistically significant increase in physical demand, thereby preserving the system's transparency.
%\textit{Physical Demand} ($F(1, 36) = 3.32, p = 0.077$)
%
Furthermore, no significant interaction effects between \textit{Task} and \textit{Condition} were observed for any of the metrics ($p > 0.05$). This indicates that the beneficial impact of \acrshort{vf}s is consistent across different levels of task complexity.
As shown in Fig.~\ref{fig:estimated_means}, in the absence of guidance, metrics such as \textit{Mental Demand} and \textit{Frustration} were notably higher and exhibited greater variability, particularly in Task 3. The activation of the \acrshort{vf} not only lowered the absolute scores but also homogenized the user experience.
Since \textit{Physical Demand} remained unchanged, the significant reduction in \textit{Effort} can be primarily attributed to the lowered cognitive burden, confirming that the virtual constraint allows users to achieve task goals with significantly lower overall exertion.}
\subsubsection{System Usability Scale}
Finally, the overall usability of the proposed assistance system was assessed via the System Usability Scale (SUS)~\cite{brooke1996sus}. This standardized questionnaire was administered \textcolor{black}{as a summative evaluation at the end of the experimental session} to provide a global measure of subjective usability (scaled 0-100), quantifying perceived ease of use and learnability.
The SUS scores averaged 74.6, exceeding the commonly accepted benchmark of 68~\cite{bangor2008empirical}. Most participants rated the system positively, with Users 5 and 6 reaching excellent levels (SUS = 85). Only User 2 scored slightly below the benchmark (SUS = 65). However, given the high consistency among the other participants and the contradictory responses within User 2's questionnaire, this case can be considered an outlier related to individual confidence rather than a reflection of system usability issues.
\begin{figure}[t]
    \centering
    \includegraphics[width=1\columnwidth]{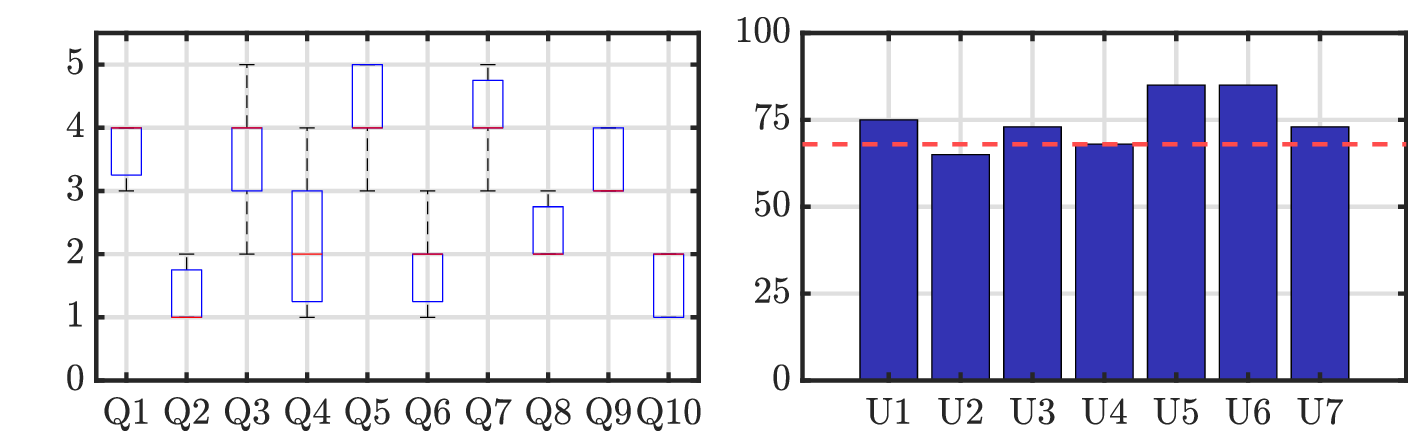}\\
    {\hspace{8mm}\small (a) \hspace{4cm}(b)}
    \caption{System Usability Scale results. (a) Questionwise distribution of scores. The items correspond to the standard SUS questions~\cite{brooke1996sus}. (b) Total SUS scores for each participant.}
    \label{fig:sus_results}
\end{figure}
\subsection{Overall Discussion}
\label{subsec:overall_discussion}
The findings presented above confirm our hypothesis and positively validate the intended efficacy of the \acrshort{emg}-based adaptive \acrlong{vf}. The alignment between objective performance metrics and subjective feedback demonstrates that the proposed assistance method significantly improves task execution compared to free motion.
Specifically, being able to utilize the guidance \acrshort{vf} leads to demonstrably lower tracking errors and higher movement consistency. By confining the tool motion within a safe corridor, the system implicitly addresses the critical safety requirement of limiting penetration depth, preventing accidental damage to underlying structures during blind cutting phases. The lower NASA-TLX scores and high SUS ratings indicate a substantial reduction in perceived workload, which could facilitate efficient tissue manipulation in complex surgical tasks.
A cross-task comparison reveals that these benefits are most pronounced in Task 3. Unlike the single-task scenarios, Task 3 imposes a cognitive overhead due to frequent switching between tracking and targeting. The fact that the \acrshort{vf} induces the highest relative improvement in this hybrid task confirms that the assistance becomes increasingly critical as dynamic demands rise.
Although the proposed application shows promising results, it is crucial to address the limitations encountered. First, the validation was conducted on a simplified drawing task rather than actual tissue dissection, which excludes factors such as tissue deformation and bleeding. Second, the pilot study involved a limited sample size of novice users. While this choice was appropriate for assessing the initial intuitiveness of the interface, evaluating the system with experienced surgeons in a realistic clinical simulation will be crucial to confirm its efficacy in professional practice.

%last update 2026/02/1
\section{Conclusion}
The pilot user study confirmed the effectiveness of this human-in-the-loop approach. The results demonstrated that the adaptive assistance significantly enhances tracking precision and movement consistency without imposing additional physical effort. Crucially, the system proved capable of mitigating the cognitive load associated with complex task-switching, validating the transparency and intuitiveness of the interface.
Future research will focus on validating these findings with expert surgeons in realistic clinical scenarios. Beyond the current explicit control, we aim to advance the system's autonomy by integrating multi-modal data (e.g., gaze tracking and instrument kinematics). This will enable the development of a context-aware controller capable of autonomously recognizing surgical phases and adapting the assistance level to the surgeon's intent, paving the way for intelligent and collaborative surgical robotics.

%\balance

\begingroup
%\footnotesize % O \small se footnote è troppo piccolo
\bibliographystyle{IEEEtran}
\bibliography{bibliography_First_Submission}
\endgroup

\end{document}